# Extraction of Medication Names from Twitter Using Augmentation and an Ensemble of Language Models

Igor Kulev[1a], Berkay Köprü[1b], Raul Rodriguez-Esteban[2], Diego Saldana[3],
Yi Huang[1c], Alessandro La Torraca[1d], Elif Ozkirimli[1a]

1. Data and Analytics Chapter, {a: F. Hoffmann-La Roche Ltd, Switzerland; b: Roche Müstahzarları San A.S., Turkey; c: Roche (China) Holding Ltd., China; d: Roche S.p.A., Italy}
2. Pharmaceutical Research and Early Development, Roche Innovation Center Basel, Switzerland
3. Personalized Healthcare Center of Excellence, F. Hoffmann-La Roche Ltd, Basel, Switzerland

*Abstract*— The BioCreative VII Track 3 challenge focused on the identification of medication names in Twitter user timelines. For our submission to this challenge, we expanded the available training data by using several data augmentation techniques. The augmented data was then used to fine-tune an ensemble of language models that had been pre-trained on general-domain Twitter content. The proposed approach outperformed the prior state-of-the-art algorithm *Kusuri* and ranked high in the competition for our selected objective function, overlapping F1 score.

## I. INTRODUCTION

We report our participation in track 3 of the BioCreative VII challenge, which involved the automatic extraction of medication names from tweets in Twitter user timelines. Closely-related research includes participant entries in task 1 of the Social Media for Health Mining (#SMM4H) 2018 Shared Tasks concerning automatic detection of posts mentioning a drug name (1). The Kusuri algorithm leveraged the findings from this challenge with the use of an ensemble approach (2). The manually-balanced corpus created for the challenge did not, however, reflect the imbalanced nature of Twitter medication-mention data. Weissenbacher et al. (3) returned to this task using a set of Twitter user timelines and focusing on strategies to deal with data imbalance. Additional related work includes participant entries in task 1 of the #SMM4H 2021 Shared Tasks, which involved the identification of adverse drug effects in tweets (4).

Much recent work in the area of named-entity recognition (NER) has been based on pre-trained language models (PLMs), which have frequently shown state-of-the-art performance. A typical approach has been to pre-train PLMs on non-specific content and fine-tune or adapt them to specific tasks. However, PLMs that have been pre-trained on tweet corpora have shown better performance in tweet-specific tasks than PLMs pre-trained on more general content (5). This is probably due to the unique structure, length and writing style of tweets compared to other types of text. Thus, one of our aims in this competition was to explore the use of PLMs that had been pre-trained specifically with tweets. Pre-training PLMs is computationally demanding, but publicly available PLMs such as TwitterBERT (6), TweetBERT (7) and BERTweet (5) are available.

Additionally, we explored ensemble strategies that integrate the output of multiple algorithms (8–9) and tested several methods to deal with data imbalance and lack of positive samples by augmenting the dataset.

## II. DATASET

The BioCreative VII Track 3 dataset consists of 127,153 tweets posted by 212 Twitter users during and after their pregnancy. Only 311 tweets in the dataset mention at least one drug name or a dietary supplement. Thus, the distribution of drug mentions is highly imbalanced. The competition organizers split the dataset into two sets with a 7:3 ratio, *training* and *validation*, such that the proportion of tweets with drug mentions in each set was preserved. The most frequently occurring drug names in these datasets are "epidural" (30 times), "tums" (16 times), "tylenol" (14 times), "birth control" (13 times), and "prenatal vitamins" (13 times). The models were evaluated on a separate *test* dataset that consists of 54,482 tweets. The goal of the BioCreative VII Track 3 Shared Task was to identify the drug names occurring in the *test* dataset and predict their correct text spans.

The organizers offered an additional dataset that could be used for training. This dataset was curated for the SMM4H'18 Task 1 (1) and consists of 9,622 tweets, out of which 4,975 mention a drug name. The most frequently occurring drug names in this dataset are "xanax" (421 times), "benadryl" (223 times), "tylenol" (220 times), "advil" (190 times), and "adderall" (180 times). The tweets in this dataset were posted by 7,586 Twitter users, a much larger number of users than in the BioCreative VII Track 3 dataset.

## III. METHOD

Our strategy in tackling the NER problem was to (i) develop models trained in different ways such that there would be higher diversity in their output, and (ii) aggregate the outputs of the individual models using an ensemble approach to generate robust predictions. Our models were based on the BERT architecture (10) and they were pre-trained on general domain Twitter data using the RoBERTa pre-training procedure (11). The first two models, *bertweet-base* and *bertweet-large* are variants of BERTweet (5). They were trained on more than 850 million English tweets. The main difference between these two models is in the number of parameters: *bertweet-large* has 355 million parameters, whereas *bertweet-base* has 135 million parameters. Another difference is that *bertweet-large* was trained on 23 million additional English tweets related to the COVID-19 pandemic. The third model, *twitter-roberta-base* (12) was trained on 58 million tweets. In contrast to BERTweet, which was trained on tweets from scratch, *twitter-roberta-base* was obtained by re-training the general-domain RoBERTa-base model (11) on Twitter data. These models also differ in the way they tokenize the data. BERTweet learns a vocabulary directly from the tweets of the corpus, whereas *twitter-roberta-base* uses the existing vocabulary from the RoBERTa-base model. In our experiments, we use these base models as implemented in the HuggingFace framework (13).

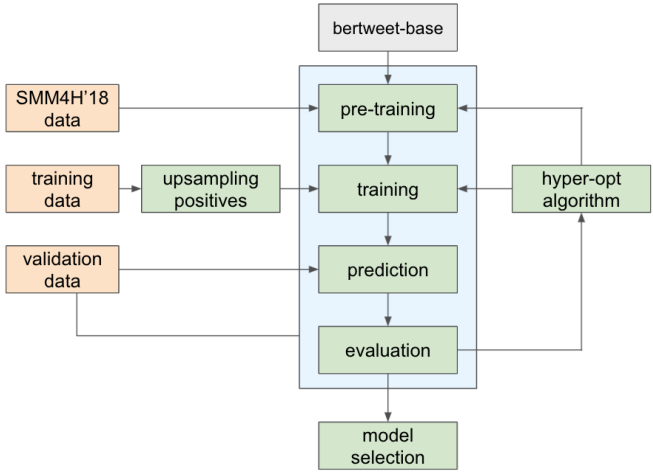 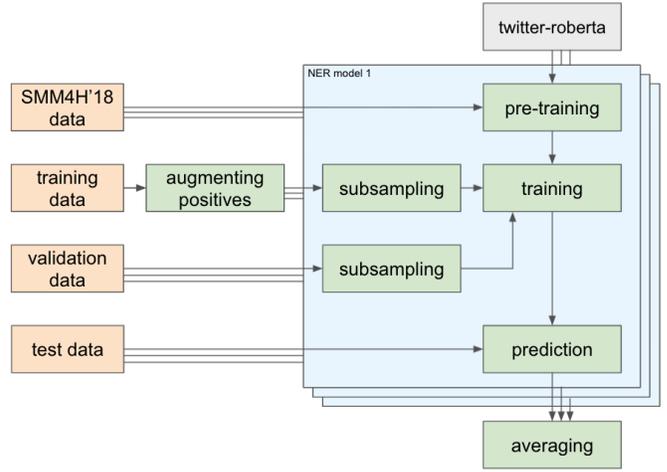

Fig. 1. The graph shows the first pipeline that we created to train a NER model. The blue rectangle represents the training procedure for a single model. The pipeline includes a hyper-parameter optimization procedure to find optimal hyper-parameters for the model.

Fig. 2. The graph shows the second pipeline that we used to train a NER model. The pipeline is an ensemble approach where each model is trained on a subset of the data (sampled with replacement). The subset from the validation data is used for early stopping.

We developed two different pipelines to fine-tune these models on the NER task using data from the BioCreative VII Track 3 challenge. In both pipelines, we pre-trained the models on the SMM4H'18 dataset. The first pipeline (PL1) used the challenge's training data augmented by upsampling tweets that contained drug names. This pipeline is illustrated in Fig. 1. We repeated the experiment with different hyper-parameters and we chose the model with the lowest cross-entropy loss on the validation dataset. We focused our search on two hyper-parameters that are very important for the learning process: the learning rate and the number of training epochs. The hyper-parameter optimization procedure used Tree of Parzen Estimators (TPE) (14) to identify promising hyper-parameter configurations. The optimal hyper-parameters were used to train a model that generates the final predictions on the test dataset. The final model was trained on the union of the training and validation data.

The second pipeline (PL2) augmented the tweets that contain drug names by (i) tweet concatenation, (ii) tweet paraphrasing, (iii) drug name replacement, and (iv) upsampling. This pipeline is illustrated in Fig. 2. The first augmentation strategy randomly picked two tweets that mention drugs and concatenated them. The second augmentation generated paraphrases of the tweets using a T5 model (15). The third augmentation replaced the drug name in a tweet with either a drug name from another tweet in the training dataset, or a drug name from a predefined manually-compiled list of drugs. The list contained drugs commonly used during pregnancy, including over-the-counter medications, prenatal vitamins, vaccines, heartburn medication, antacids, NSAIDs, pain relief medication, among others. After the augmentation phase, the pipeline created six different subsets of the training and the validation datasets by subsampling with replacement. Afterwards, these subsamples were used to train six models. These models were based on the same architecture and hyper-parameters but trained on randomly-selected subsets of the data. All model predictions were aggregated by averaging. By repeating the training on subsets of the data, we aimed to obtain a more robust model that is less overfitted to the training data.

We ran the first pipeline twice, each time using a different version of BERTweet (5) as a base model. We refer to the resulting models as *PL1-bertweet-base* and *PL1-bertweet-large*. We ran the second pipeline once using Twitter-RoBERTa as a base model and we refer to the resulting model as *PL2-twitter-roberta*. Thus, we obtained three models with differences in architecture, data, or training procedure used. We used an *ensemble* approach to make the final predictions on the test data based on the predictions from the three individual models by doing a character-level weighted sum of the models' outputs and setting a predefined threshold to assign them to a drug name prediction. We treated model weights and the threshold as hyper-parameters and we used both grid-search and TPE optimization to find the optimal values. Our objective function was the overlapping F1 score on the validation dataset.

IV. RESULTS

Both grid-search and TPE optimization converged to the same optimal weights for the ensemble approach. We observed that a character was considered to be a part of a drug name if at

TABLE I. OVERLAPPING AND STRICT PRECISION, RECALL AND F1 SCORE ON THE VALIDATION SET

| Model | Overlapping | | | Strict | | |
|---|---|---|---|---|---|---|
| | F1 score | Precision | Recall | F1 score | Precision | Recall |
| PL1-bertweet-base | 0.856 | 0.836 | **0.876** | 0.804 | 0.789 | **0.819** |
| PL1-bertweet-large | 0.847 | **0.912** | 0.790 | 0.790 | 0.856 | 0.733 |
| PL2-twitter-roberta-base | 0.834 | 0.830 | 0.838 | 0.815 | 0.811 | **0.819** |
| ensemble | **0.870** | 0.882 | 0.857 | 0.816 | 0.832 | 0.800 |
| ensemble-post | **0.870** | 0.882 | 0.857 | **0.825** | 0.842 | 0.810 |

TABLE II. OVERLAPPING AND STRICT PRECISION, RECALL AND F1 SCORE ON THE TEST DATASET

| Model | Overlapping | | | Strict | | |
|---|---|---|---|---|---|---|
| | F1 score | Precision | Recall | F1 score | Precision | Recall |
| kusuri-CNN | 0.717 | 0.769 | 0.671 | 0.652 | 0.700 | 0.611 |
| kusuri-BERT | 0.773 | **0.908** | 0.673 | **0.758** | **0.890** | 0.660 |
| avg. of task entries[a] | 0.749 | 0.811 | 0.709 | 0.696 | 0.754 | 0.658 |
| PL1-bertweet-base | **0.804** | 0.812 | 0.796 | 0.727 | 0.739 | 0.714 |
| PL1-bertweet-large | 0.801 | 0.827 | 0.777 | 0.697 | 0.723 | 0.673 |
| PL2-twitter-roberta-base | 0.710 | 0.720 | 0.701 | 0.683 | 0.692 | 0.673 |
| ensemble | **0.804** | 0.827 | **0.782** | 0.725 | 0.752 | 0.701 |
| ensemble-post | **0.804** | 0.827 | **0.782** | 0.754 | 0.781 | **0.728** |

[a.] The statistics on the best submissions for all 16 participants have been provided by the competition organizers

least two out of the three models voted "yes." This showed that all three models were equally important to generate final predictions that were better than the predictions from individual models. Our submission to the BioCreative VII Track 3 challenge was generated using the *ensemble* model. After the competition ended we added a post-processing step to exclude hashtag symbols from the drug spans predictions generated by the ensemble model, which had produced errors affecting the strict F1 score. We refer to this model as *ensemble-post*.

The model performance was evaluated in two ways. The *strict* F1 score, precision and recall measure how well the predictions match the correct spans of the drug names. The *overlapping* F1 score, precision and recall measure how well the predictions overlap with the correct spans of the drug names. In Table I, we report the performance of our models on the validation dataset. We note that the validation data was used for hyper-parameter optimization, as illustrated in Fig. 1, and the model performance was evaluated based on this dataset. Therefore, the results reported in Table I may be optimistic due to overtraining or inability to generalize.

The F1 scores of the three models that were included in our *ensemble* approach were similar and ranged from 0.834 to 0.856 in the overlapping case and from 0.790 to 0.815 in the strict case. The ensemble model outperformed all three individual models in terms of F1 score and therefore we selected it to generate the final predictions on the test dataset. We manually inspected the predictions from our final model and observed that some of the drug mentions predicted did not appear in the SMM4H'18 Task 1 dataset nor in the BioCreative VII Track 3 training dataset. This means that our model was able to identify drugs based on the context and with robustness towards typographical errors, e.g., "asprin" instead of "aspirin".

The performance of our ensemble model on the test dataset is given in Table II. We compare our models with different versions of the Kusuri model as described in (3). Kusuri performs drug extraction using a Convolutional Neural Network (CNN) or a BERT model. We refer to these models as *kusuri-CNN* and *kusuri-BERT*, respectively. They were trained on the SMM4H'18 training set and fine-tuned on the BioCreative VII Track 3 training set. We additionally compare our models with the average performance of the 16 challenge submissions (*avg. of task entries*), which, together with the standard deviation, was the only overall challenge information provided by the organizers at the time of this writing. The results show that our model performs better than the average submission in all six metrics and it performs better than the Kusuri models in terms of overlapping F1 score. The results on the strict metrics were further improved by performing hashtag removal (*ensemble-post*). For example, the strict F1 score increased from 0.725 (the *ensemble* model) to 0.754 (the *ensemble-post* model). In this way, the overlapping and strict F1 scores from *ensemble-post* were approximately one standard deviation above the mean scores of the 16 submissions. The strict F1 score for the *ensemble-post* model was lower than the one for *kusuri-BERT*, but similar.

TABLE III. ERROR ANALYSIS OF THE ENSEMBLE CLASSIFIER OUTPUT ON THE VALIDATION SET

| Error category | #Errors | Examples |
|---|---|---|
| *False positive types* | | |
| Annotation error | 11 | Prenatal vitamins, birth control, flu shot |
| *False negative types* | | |
| Drug not/rarely seen | 6 | Bio-oil, blood patch |
| Other | 6 | Vicks vapo rub |
| Complex drug phrase | 2 | #LifeWithAZofranPump |
| Annotation error | 1 | I'm in charge of narcotics & domestic violence |

We analyzed the labeling errors made by the *ensemble* model on the validation set and noted that a majority of false positives were a result of annotation errors (see Table III). For example, the term "birth control" was annotated as a drug in both the SMM4H'18 dataset and the BioCreative VII Track 3 training dataset, but not in the BioCreative VII Track 3 test dataset. We distinguished four non-exclusive categories of false negatives. Most of these errors were probably caused by insufficient training data on the drugs involved. We also manually inspected the cases where the model recognized the drug but did not predict the span correctly. We observed that often the span included non-alphabetic characters, for example our model predicted "Follistim:" instead of "Follistim", suggesting that our performance on the strict metrics could have been higher by improving either the tokenization or the post-processing procedure.

## V. CONCLUSION

Processing Twitter data is not straightforward for two main reasons: the data is highly imbalanced with many more negative instances than positive instances, and the text is highly unique with a particular structure, style, length, and vocabulary. Therefore, it is a challenge to find the proverbial needles in the haystack with common language models. In this work, we have tried to address the class imbalance problem by data augmentation, and the unique Twitter language characteristics by using three language models trained on tweets. Finally, we ensembled the predictions to take advantage of the strengths of different approaches. Our selected objective function during the competition was the overlapping F1 score and our results with this metric, both in *ensemble* and *ensemble-post*, ranked high compared to competing entries. Future avenues of exploration could include the addition of further knowledge to the models, such as lexicons and patient-specific language, or timeline-specific temporal and sequential information.